\documentclass{article} 
\usepackage{colm2024_conference}

\usepackage{microtype}
\usepackage{hyperref}
\usepackage{url}
\usepackage{times}
\usepackage{latexsym}
\usepackage{color}
\usepackage{pgfplots}
\pgfplotsset{compat=1.18}
\usepackage{xcolor}
\usepackage{tikz}
\usepackage{placeins}
\usepackage{amsmath}
\usepackage{float}
\usepackage{algorithm}      
\usepackage{algpseudocode}
\usepackage{amsmath}

\definecolor{baselinegray}{RGB}{60, 60, 60}
\definecolor{influencecolor}{RGB}{102, 194, 165}   
\definecolor{gradcolor}{RGB}{252, 141, 98}        
\definecolor{randomcolor}{RGB}{141, 160, 203}      
\usepackage{booktabs}
\definecolor{darkblue}{rgb}{0, 0, 0.5}
\hypersetup{colorlinks=true, citecolor=darkblue, linkcolor=darkblue, urlcolor=darkblue}

\title{Influence Functions for Preference Dataset Pruning}

\author{Daniel Fein \& Gabriela Aranguiz-Dias\\
Department of Computer Science\\
Stanford University\\ \\
\texttt{\{drfein, gadias\}@stanford.edu} \\
}

\colmfinalcopy 
\begin{document}

\maketitle

\begin{abstract}
Language models are commonly fine-tuned via reinforcement learning to alter their behavior or elicit new capabilities. Datasets used for these purposes, and particularly human preference datasets, are often noisy. The relatively small size post-training datasets, combined with parameter-efficient fine-tuning methods, enable the use of influence functions approximations to detect and prune training examples that are harmful to performance on a validation set. In this work, we adapt the TL;DR dataset for reward model training to demonstrate how conjugate-gradient approximated influence functions can be used to filter datasets. In our experiments, influence function filtering yields a small retraining accuracy uplift of 1.5\% after removing 10\% of training examples. We also show that gradient similarity outperforms influence functions for detecting helpful training examples. This suggests that local curvature is important for detecting harmful training examples, but less so for identifying helpful examples.
\end{abstract}

\section{Introduction}
Influence functions are powerful tools for understanding how specific data affect the behavior of neural networks trained via gradient descent \citep{koh2017understanding}. However, their practicality for language models is encumbered by their reliance on the inverse-hessian with respect to model parameters. The large scale of language models, both in terms of parameter count and training data, thus poses a challenge for influence-function based data attribution on the scale of pre-training. Despite this, approximations of the inverse hessian used at scale have demonstrated the power of influence functions to reveal insights into how language models make use of their training data \cite{grosse2023studyinglargelanguagemodel}.

Recently, the post-training paradigm has proven crucial for controlling the behavior of language models both with respect to helpfulness and harmlessness as well as math and coding abilities \citep{bai2022training, zelikman2022star}. Meanwhile, techniques for parameter-efficient fine-tuning such as Low-Rank Adaptation enable this post-training to be done via training of relatively few model parameters \cite{hu2022lora}. The outsized effectiveness of this much less computationally-expensive training warrants a new investigation into influence functions for understanding the relationship between data and model behavior.

Turning specifically to the task of human preference modeling, it is estimated that 20-40\% of training data used to align language models with human preferences is noisy \citep{gao2024impactpreferencenoisealignment}. Influence functions have been used to reduce the size of training sets for computer vision tasks \citep{yang2023datasetpruningreducingtraining}, but have not yet been applied to language model data curation.

In this work, we demonstrate for the first time how influence functions approximated via the conjugate gradient method can help create stronger reward models via filtering out of noisy examples.

\section{Related Works}
\textbf{Influence Function Approximation}
\cite{martens2010deep} first approximated the hessian of deep networks using the conjugate gradient method,and \cite{koh2017understanding} applied this to influence functions. \cite{kwon2023datainf} and \cite{grosse2023studyinglargelanguagemodel} propose algorithms that efficiently approximate the conjugate gradient method, and apply them to LoRA-finetuned and pretrained language models, respectively. They do not evaluate the conjugate gradient method for data curation, nor do they consider the task of preference modeling. 

\textbf{RLHF and Data Curation}
\cite{stienon2020learning} showed that a language model could be taught to generate better summaries using human feedback in the form of pairwise preferences. We use the TL;DR dataset they provide. \cite{bai2022traininghelpfulharmlessassistant} post-trained a language model to align it's behavior to be helpful and harmless. \cite{morimura2024filtereddirectpreferenceoptimization} and \cite{gao2024impactpreferencenoisealignment} point out and attempt to address low data quality for preference data. The former does this by proposing a framework for data collection, the latter by proposing filtering based on trained reward-model logits.

\section{Methods}
\label{sec:methods}

\subsection{Task and Data}
\label{subsec:data}
We study the human–preference reward–modeling task introduced by \citet{bai2022training} on the public \textsc{TL;DR} dataset released by OpenAI.\footnote{\url{https://github.com/openai/summarize-from-feedback}}  
Each training example is a \textit{chosen} and a \textit{rejected} summary of the same news article, labeled by crowd workers.  
Following common practice, we cast the task as a Bradley–Terry pairwise–ranking problem: the reward model $f_\theta$ should assign a higher scalar reward to the chosen summary than to its rejected counterpart.

\vspace{2pt}\noindent
\textbf{Dataset Modification}.  
In order to make the problem tractable under a very tight computation budget, we filter out all examples for which both summaries are longer than 24 tokens Llama3 tokens, and discard the original post text, using only summary preferences and summaries themselves. This leaves us with roughly 8.5k examples total. 
For influence estimation we sample $|\mathcal D_\text{val}| = 100$ validation pairs uniformly at random from the held‑out validation set.  
All remaining validation data serve as an untouched test set for final evaluation.

\subsection{Base Model and Fine‑tuning Setup}
\label{subsec:finetune}
Our base model is \textbf{LLaMA-3.2‑1B}.  
We apply Low‑Rank Adaptation (\textsc{LoRA}; \citealp{hu2022lora}) with rank $r{=}8$, $\alpha{=}16$ and dropout $0.05$ to the projection matrices
$\{q,k,v,o\}$.  
Fine‑tuning uses the AdamW optimiser ($\beta_1{=}0.9,\ \beta_2{=}0.98,\ \epsilon{=}10^{-8}$) for three epochs, batch size $124$ and learning rate $1\times10^{-5}$ with cosine decay.  
Only LoRA parameters ($\approx0.12\%$ of total weights) are updated.

\subsection{Influence–Function Approximation}
\label{subsec:influence}

The goal is to estimate, for every training example $z_i$ and validation
example $z_j$, the classical \emph{influence} value
$\mathcal I_{\text{IF}}(z_i,z_j)=
     -\nabla_\theta L(z_j)^\top H_\theta^{-1}\nabla_\theta L(z_i)$,
where $H_\theta$ is the Hessian of the total training loss.
Below we give an intuitive walk‑through of the procedure, then collect the
steps into Algorithm~\ref{alg:influence-lora}. Instead of inverting a billion‑dimensional Hessian, we study only the
\textbf{LoRA adapter weights}.  These account for $0.12\%$ of all parameters
($\approx\!1.2$\,M scalars) yet dominate post‑training behavior
\citep{hu2022lora}.  All gradients are therefore taken with respect to this small sub‑space, making both
memory use and matrix‑vector multiplies tractable.

\textbf{The Optimization Perspective}
Computing $H_\theta^{-1}\nabla_\theta L(z_j)$ is equivalent to solving the
linear system
\[
  (H_\theta + \lambda I)\,x = g_j
  \quad\text{with}\quad g_j := \nabla_\theta L(z_j),
\]
where we add Tikhonov damping $\lambda{=}10^{-2}$ to guarantee positive
definiteness.  This linear system is the first‑order optimality condition of
the strictly convex quadratic
\[
  \min_{x\in \mathbb R^d}\;
    \frac12\,x^\top (H_\theta+\lambda I)x \;-\; g_j^\top x,
\]
so the problem can be viewed as an optimization problem whose solution $x^\star$
is the desired inverse‑Hessian–vector product. To computing an HVP without forming $H_\theta$, we use the \textit{double‑back‑prop} trick
\citep{pearlmutter1994fast}.

We average the
operation over a mini‑batch of $|\mathcal B|=20$ training examples,
reducing variance while staying limiting GPU memory use.

\begin{algorithm*}[t]                  
  \caption{Influence computation in the LoRA parameter space}
  \label{alg:influence-lora}

  \begin{algorithmic}[1]             
    \Require
      LoRA‑tuned model $f_\theta$; training set $\mathcal{D}_{\text{train}}$;
      validation set $\mathcal{D}_{\text{val}}$; damping $\lambda$;
      batch size $B$; CG iteration budget $K$.

    \ForAll{$z_j \in \mathcal{D}_{\text{val}}$}          \Comment{loop over validation pairs}
      \State $g_j \gets \nabla_\theta L(z_j)$            \Comment{retain comp.\ graph}
      \State $x_j \gets 0$; \ $r \gets g_j$; \ $p \gets r$
      \For{$k = 1$ \textbf{to} $K$}                      \Comment{conjugate‑gradient iterations}
        \State $\mathcal{B} \gets$ sample $B$ elements from $\mathcal{D}_{\text{train}}$
        \State $h \gets \dfrac{1}{\lvert\mathcal{B}\rvert}
                     \sum_{z_i \in \mathcal{B}}
                     \nabla_\theta^{2}L(z_i)\,p \;+\; \lambda p$
                     \Comment{batched HVP via double back‑prop}
        \State $\alpha \gets \dfrac{\langle r,r\rangle}{\langle p,h\rangle}$
        \State $x_j \gets x_j + \alpha p$
        \State $r_{\text{new}} \gets r - \alpha h$
        \If{$\lVert r_{\text{new}}\rVert < \varepsilon$}
          \State \textbf{break}
        \EndIf
        \State $\beta \gets \dfrac{\langle r_{\text{new}},\,r_{\text{new}}\rangle}
                             {\langle r,\,r\rangle}$
        \State $p \gets r_{\text{new}} + \beta p$
        \State $r \gets r_{\text{new}}$
      \EndFor
      \State $x_j \gets \textbf{detach}(x_j)$            \Comment{$x_j \approx H_\theta^{-1}g_j$}
      \ForAll{$z_i \in \textbf{local\_shard}(\mathcal{D}_{\text{train}})$}
        \State $g_i \gets \nabla_\theta L(z_i)$
        \State $\mathcal{I}(z_i,z_j) \gets -\langle x_j,\,g_i\rangle$   \Comment{influence score}
      \EndFor
    \EndFor
    \State \Return $\mathcal{I}$ collected on rank 0
  \end{algorithmic}
\end{algorithm*}

\subsection{Filtering Strategy}
\label{subsec:filter}
Let $\bar{\mathcal I}(z_i)=\frac1{|\mathcal D_\text{val}|}\sum_{z_j\in\mathcal D_\text{val}}\mathcal I_{\text{IF}}(z_i,z_j)$ denote the mean influence of a training example across validation points.  
We \textbf{rank} all $z_i\in\mathcal D_\text{train}$ by $\bar{\mathcal I}(z_i)$ and drop the worst $x\%$ (most positive) examples. 

\subsection{Retraining and Evaluation}
After pruning, we \emph{retrain the reward model from the original checkpoint} under the identical hyper‑parameter schedule of Section~\ref{subsec:finetune}.  
We report pairwise accuracy on the untouched test split.

\subsection{Baselines}
We use two baselines to measure the effectiveness of this method for data curation. The first is a random baseline where we randomly remove $x\%$ of training examples. Our second baseline is first-order influence approximation that presumes the Hessian to be the identity, also known as gradient similarity (\cite{dhaliwal2018gradientsimilarityexplainableapproach}).

\section{Results}
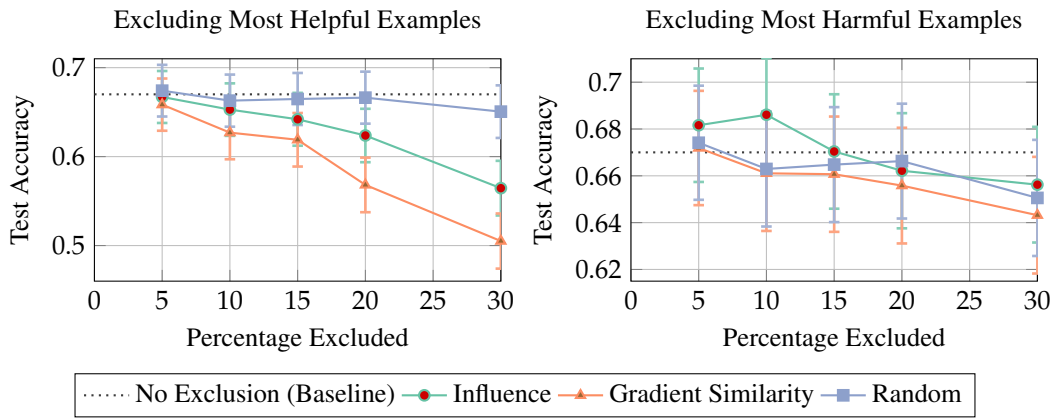
\begin{figure*}[t]
  \centering
  \begin{minipage}[t]{0.5\textwidth}
    \centering
    \begin{tikzpicture}
      \begin{axis}[
        title={Excluding Most Helpful Examples},
        xlabel={Percentage Excluded},
        ylabel={Test Accuracy},
        grid=major,
        xmin=0,xmax=30,
        ymin=0.46,ymax=0.71,
        xtick={0,5,10,15,20,25,30},
        yticklabel style={/pgf/number format/.cd,fixed,precision=3},
        width=\linewidth,height=0.65\linewidth,
        mark size=2pt,
      ]
        \addplot[baselinegray,dotted,thick]
                 coordinates {(0,0.67002) (30,0.67002)};

        \addplot+[influencecolor,mark=*,thick,
            error bars/.cd,
              y dir=both, y explicit,
              error bar style={influencecolor!80, line width=1pt},
              error mark=|,
              error mark options={influencecolor!80, line width=1pt}]
          coordinates {
            (5, 0.667040) +- (0,0.02921)
            (10,0.652855) +- (0,0.02951)
            (15,0.641975) +- (0,0.02971)
            (20,0.623740) +- (0,0.03003)
            (30,0.564534) +- (0,0.03073)
          };

        \addplot+[gradcolor,mark=triangle*,thick,
            error bars/.cd,
              y dir=both, y explicit,
              error bar style={gradcolor!80, line width=1pt},
              error mark=|,
              error mark options={gradcolor!80, line width=1pt}]
          coordinates {
            (5, 0.658454) +- (0,0.02936)
            (10,0.626966) +- (0,0.02994)
            (15,0.618940) +- (0,0.03010)
            (20,0.568200) +- (0,0.03070)
            (30,0.505073) +- (0,0.03099)
          };

        \addplot+[randomcolor,mark=square*,thick,
            error bars/.cd,
              y dir=both, y explicit,
              error bar style={randomcolor!80, line width=1pt},
              error mark=|,
              error mark options={randomcolor!80, line width=1pt}]
          coordinates {
            (5, 0.674132) +- (0,0.02905)
            (10,0.662933) +- (0,0.02930)
            (15,0.664790) +- (0,0.02926)
            (20,0.666290) +- (0,0.02923)
            (30,0.650561) +- (0,0.02955)
          };
      \end{axis}
    \end{tikzpicture}
  \end{minipage}\hfill
  \begin{minipage}[t]{0.5\textwidth}
    \centering
    \begin{tikzpicture}
      \begin{axis}[
        title={Excluding Most Harmful Examples},
        xlabel={Percentage Excluded},
        ylabel={Test Accuracy},
        grid=major,
        xmin=0,xmax=30,
        ymin=0.615,ymax=0.71,
        xtick={0,5,10,15,20,25,30},
        yticklabel style={/pgf/number format/.cd,fixed,precision=3},
        width=\linewidth,height=0.65\linewidth,
        mark size=2pt,
        legend columns=4,
        legend to name=shared,
      ]
        \addplot[baselinegray,dotted,thick]
                 coordinates {(0,0.67002) (30,0.67002)};
        \addlegendentry{No Exclusion (Baseline)}

        \addplot+[influencecolor,mark=*,thick,
            error bars/.cd,
              y dir=both, y explicit,
              error bar style={influencecolor!80, line width=1pt},
              error mark=|,
              error mark options={influencecolor!80, line width=1pt}]
          coordinates {
            (5, 0.681610) +- (0,0.02423)
            (10,0.686090) +- (0,0.02414)
            (15,0.670400) +- (0,0.02445)
            (20,0.662170) +- (0,0.02460)
            (30,0.656215) +- (0,0.02471)
          };
        \addlegendentry{Influence}

        \addplot+[gradcolor,mark=triangle*,thick,
            error bars/.cd,
              y dir=both, y explicit,
              error bar style={gradcolor!80, line width=1pt},
              error mark=|,
              error mark options={gradcolor!80, line width=1pt}]
          coordinates {
            (5, 0.671890) +- (0,0.02442)
            (10,0.661070) +- (0,0.02462)
            (15,0.660694) +- (0,0.02463)
            (20,0.655820) +- (0,0.02471)
            (30,0.643177) +- (0,0.02492)
          };
        \addlegendentry{Gradient Similarity}

        \addplot+[randomcolor,mark=square*,thick,
            error bars/.cd,
              y dir=both, y explicit,
              error bar style={randomcolor!80, line width=1pt},
              error mark=|,
              error mark options={randomcolor!80, line width=1pt}]
          coordinates {
            (5, 0.674132) +- (0,0.02438)
            (10,0.662933) +- (0,0.02459)
            (15,0.664790) +- (0,0.02456)
            (20,0.666290) +- (0,0.02453)
            (30,0.650561) +- (0,0.02480)
          };
        \addlegendentry{Random}
      \end{axis}
    \end{tikzpicture}
  \end{minipage}

  \vspace{0.4em}
  \pgfplotslegendfromname{shared}

  \caption{\textbf{Test accuracy after excluding training examples.}
           Left: removing the most helpful examples. Right: removing the most harmful ones.
           Vertical bars show 95\,\% Wald Interval.}
  \label{fig:unified-exclusion-strategies}
\end{figure*}

Figure \ref{fig:unified-exclusion-strategies} shows that influence-based filtering of the modified TL;DR dataset is able to improve performance a maximum of 1.5\% above finetuning on the entire dataset when 10\% of examples are pruned. For this particular dataset, accuracy gains over full training are not statistically significant (test set n=1000). However, we find that influence-pruning does improve accuracy above other pruning techniques at 10\% exclusion. Further, influence-based filtering performs about the same-- very slightly better-- than the baseline, while using 15\% less data. Gradient similarity appears to perform similarly to random pruning in terms of identifying the most harmful examples.

\begin{figure*}[tb!]
    \centering
    \includegraphics[width=1\linewidth]{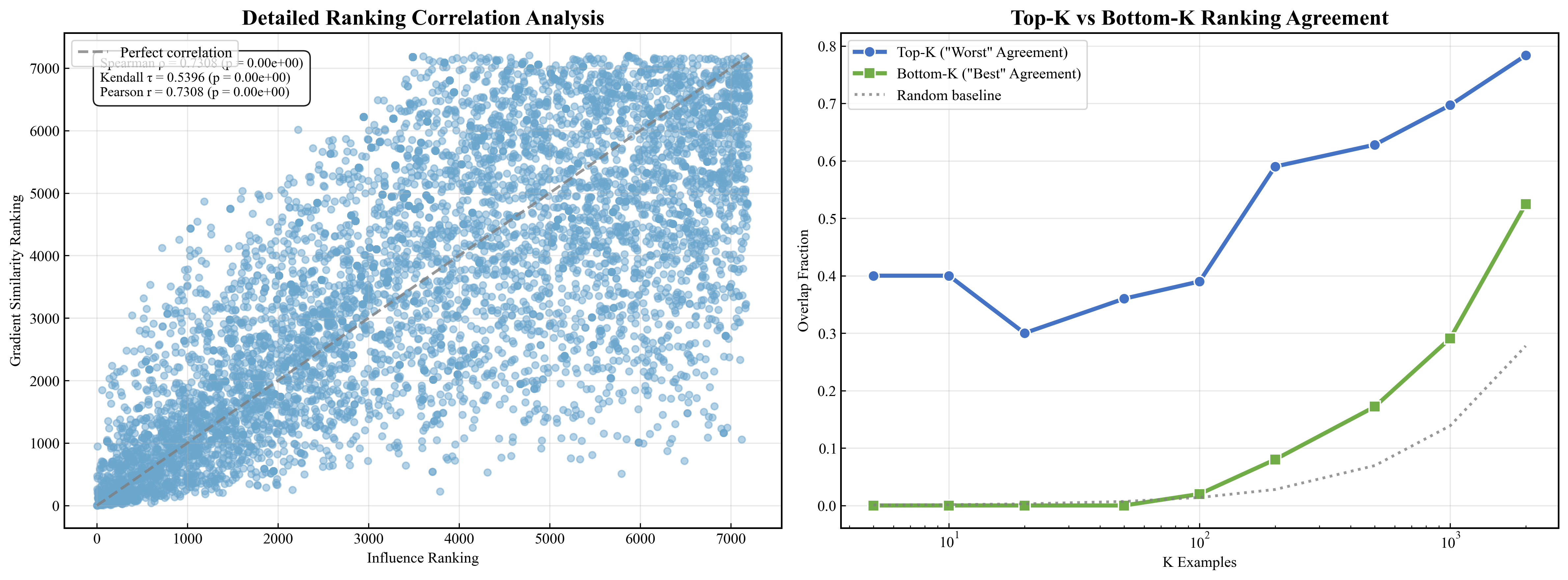}
    \caption{Left: Rank correlation between Influence and Gradient Similarity Rankings (lower ranks represent worse examples). Right: Ranking agreement between gradient similarity and influence function approaches for top/bottom k examples.}
    \label{fig:ranks}
\end{figure*}

Interestingly, we find that gradient similarity outperforms influence function approximation for finding the most helpful examples. Removing 30\% of the best training examples identified by gradient similarity reduces performance of the trained reward model to random chance, while influence based pruning only reduces performance to roughly 56\%. Figure \ref{fig:ranks} shows that there is strong agreement among what are deemed to be the worst examples, but only near random chance agreement in what is deemed to be the best examples.

\section{Discussion}
Here, we speculate that the failure of gradient similarity on finding harmful examples has to do with the more pronounced local curvature around these harmful examples. Helpful examples are those which give the model new information about a particular context. This corresponds to relatively a flat part of parameter space where not much has been learned, and therefore gradient similarity and influence functions both work well (since the inverse hessian is close to the identity). Harmful examples, on the other hand, are likely those which sit on steep inclines in the direction that go against the consensus found in the training data. Though the gradient similarity approach can determine when a given training data is poorly suited for a given validation example, it cannot determine the extent to which this error is surprising, or merely noise. This finding suggests further work that might be done to more efficiently find harmful examples to advance data curation at scale. 

\section{Conclusion}
We show that influence function approximations offer a promising approach to filtering noisy human preference data for reward model training. Further, we show that first-degree approximation via gradient similarity is more effective at finding helpful training examples, but possibly less effective at finding harmful ones. Future work may attempt to use the insight that training data that disagrees with validation data in a way that is surprising are most likely to be harming test performance to devise more efficient methods of data curation.

\section{Limitations}
Calculating influence function approximations is computationally expensive, even at smaller scales. Thus, we are constrained to using a single small dataset. Datasets are often generated in vastly different ways, and likely have highly variable noise. This limits the generalization of these results beyond the context of crowd-sourced human preference data.
\FloatBarrier

\bibliography{colm2024_conference}
\bibliographystyle{colm2024_conference}


\end{document}